\documentclass[10pt,twocolumn,letterpaper]{article}

\usepackage{iccv}
\usepackage{times}
\usepackage{epsfig}
\usepackage{graphicx}
\usepackage{amsmath}
\usepackage{amssymb}
\usepackage{multirow}
\usepackage{booktabs}
\usepackage{bbold}

\usepackage{caption}
\usepackage{subcaption}
\usepackage[accsupp]{axessibility} 

\usepackage[pagebackref=true,breaklinks=true,letterpaper=true,colorlinks,bookmarks=false]{hyperref}
\usepackage[capitalize]{cleveref}
\crefname{section}{Sec.}{Secs.}
\Crefname{section}{Section}{Sections}
\Crefname{table}{Table}{Tables}
\crefname{table}{Tab.}{Tabs.}
\Crefname{appsec}{Appendix}{Appendices}
\crefname{appsec}{App.}{Apps.}
\iccvfinalcopy 

\def\iccvPaperID{***} 

\ificcvfinal\fi

\begin{document}

\title{ICICLE: Interpretable Class Incremental Continual Learning}

\author{Dawid Rymarczyk$^{1, 2, 3, *}$
\and
Joost van de Weijer$^{4,5}$
\and
Bartosz Zieliński$^{1, 3, 6}$
\and 
Bartłomiej Twardowski$^{4,5,6}$
\and
$^1$ Faculty of Mathematics and Computer Science, Jagiellonian University
\and
$^2$ Doctoral School of Exact and Life Sciences, Jagiellonian University
\and
$^3$ Ardigen SA
\and 
$^4$ Autonomous University of Barcelona 
\and 
$^5$ Computer Vision Center
\and
$^6$ IDEAS NCBR
\and 
$^*${\tt\small dawid.rymarczyk@doctoral.uj.edu.pl}
}

\maketitle
\ificcvfinal\thispagestyle{empty}\fi

\begin{abstract}

Continual learning enables incremental learning of new tasks without forgetting those previously learned, resulting in positive knowledge transfer that can enhance performance on both new and old tasks. However, continual learning poses new challenges for interpretability, as the rationale behind model predictions may change over time, leading to interpretability concept drift.
We address this problem by proposing Interpretable Class-InCremental LEarning (ICICLE), an exemplar-free approach that adopts a prototypical part-based approach. It consists of three crucial novelties: interpretability regularization that distills previously learned concepts while preserving user-friendly positive reasoning; proximity-based prototype initialization strategy dedicated to the fine-grained setting; and task-recency bias compensation devoted to prototypical parts.
Our experimental results demonstrate that ICICLE reduces the interpretability concept drift and outperforms the existing exemplar-free methods of common class-incremental learning when applied to concept-based models. 
\end{abstract}


\vspace{-2em}
\section{Introduction}

\begin{figure}[t]
\centering
\includegraphics[width=0.8\linewidth]{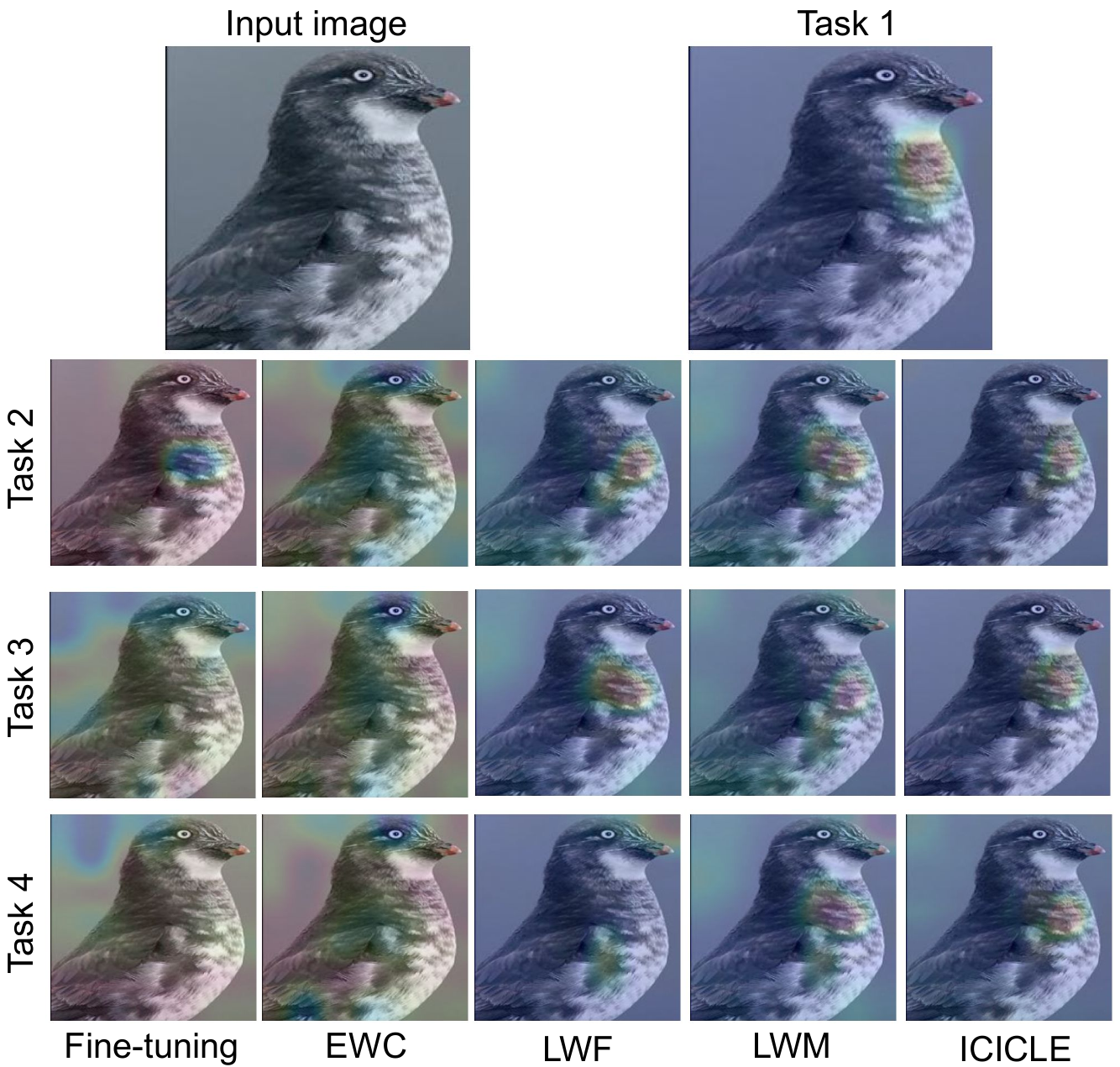}
\caption{We process the input image (top left) through the network and visualize how its specific areas are similar to one of the prototypes. The interpretability concept drift occurs when such a similarity map differs between tasks. ICICLE performs best, preserving similarity maps better than the other continual learning methods.}
\label{fig:fig1}
\vspace{-5ex}
\end{figure}

With the growing use of deep learning models in diverse domains, including robotics~\cite{brunke2022safe}, medical imaging~\cite{chen2022recent}, and autonomous driving~\cite{kiran2021deep}, there is a pressing need to develop models that can adapt to ever-changing conditions and learn new tasks from non-stationary data. However, a significant challenge with neural networks is their tendency to suffer from \emph{catastrophic forgetting}~\cite{1999french,goodfellow2013empirical,kirkpatrick2017overcoming}, where  performance on previous tasks deteriorates rapidly as new ones are acquired. Continual Learning (CL)~\cite{delange2021continual} has emerged as a promising technique to address this challenge by enabling models to learn new tasks without forgetting those learned before.

While existing CL approaches significantly reduce catastrophic forgetting, they are often difficult for humans to understand. It is especially problematic because deep networks often predict the right answer for the wrong reason (the ``Clever Hans'' phenomenon), leading to excellent performance in training but poor performance in practice~\cite{schramowski2020making}. This results in serious societal problems that deeply affect health, freedom, racial bias, and safety~\cite{rudin2022interpretable}. As a result, some initial steps were taken in the literature to introduce explainable posthoc methods into the CL setup~\cite{guzy2021evaluating,marconato2022catastrophic,patra2020incremental}. However, explaining black boxes, rather than replacing them with interpretable (self-explainable) models, can escalate the problem by providing misleading or false characterizations~\cite{rudin2019stop} or adding unnecessary authority to the black box~\cite{rudin2019we}. Therefore, there is a clear need for innovative machine-learning models that are inherently interpretable~\cite{rudin2022interpretable}. To the best of our knowledge, no interpretable CL approach has been proposed so far.

In this work, we introduce Interpretable Class-Incremental Learning (ICICLE), an interpretable approach to class-incremental learning based on prototypical parts methodology. Similarly to \emph{This looks like that} reasoning~\cite{chen2019looks}, ICICLE learns a set of prototypical parts representing reference concepts derived from the training data and makes predictions by comparing the input image parts to the learned prototypes. However, the knowledge transfer between tasks in continual learning poses new challenges for interpretability. Mainly because the rationale behind model predictions may change over time, leading to \emph{interpretability concept drift} and making explanations inconsistent (see \Cref{fig:fig1} and \Cref{table:ious}). Therefore, ICICLE contains multiple mechanisms to prevent this drift and, at the same time, obtain satisfactory results.
First, we propose an interpretability regularization suited for prototypical part-based models to retain previously gained knowledge while maintaining model plasticity. It ensures that previously learned prototypical parts are similarly activated within the current task data, which makes explanations consistent over time.
Moreover, considering the fine-grained nature of considered datasets, we introduce proximity-based prototype initialization for a new task. It searches for representative concepts within the new task data close to previously learned concepts, allowing the model to recognize high-level features of the new task and focusing on tuning details.
Thirdly, to overcome task-recency bias in class-incremental learning scenarios, we propose a simple yet effective method that balances the logits of all tasks based on the last task data.
Finally, we reduce multi-stage training while preserving user-friendly positive reasoning.

We evaluate ICICLE on two datasets, namely CUB-200-2011~\cite{wah2011caltech} and Stanford Cars~\cite{krause20133d}, and conduct exhaustive ablations to demonstrate the effectiveness of our approach. 
We show that this problem is challenging but opens up a promising new area of research that can further advance our understanding of CL methods. Our contributions can be summarized as follows:
\begin{itemize}
\vspace{-1em}
    \item We are the first to introduce interpretable class-incremental learning and propose a new method ICICLE, based on prototypical part methodology.
    \vspace{-1em}
    \item We propose interpretability regularization that prevents interpretability concept drift without using exemplars.
    \vspace{-1em}
    \item We define a dedicated prototype initialization strategy and a method compensating for task-recency bias.
\end{itemize}
\vspace{-1em}

\begin{table}[t]

\centering
\scriptsize 
\begin{tabular}{|c||c|c|c|c|}
\hline
& \multicolumn{4}{c|}{\textsc{IoU}} \\ [0.5ex]
\cline{2-5}
\textsc{Method} & \textsc{task 1} & \textsc{task 2} & \textsc{task 3} & \textsc{mean} \\ [0.5ex]
\hline
\hline
\textsc{Finetuning} & $0.115$ & $0.149$  & $0.260$ &  $0.151 $ \\ [0.5ex]
\textsc{EWC} & $0.192$ & $0.481 $  & $0.467$ &  $0.334$ \\  [0.5ex]
\textsc{LWF} & $0.221$ & $0.193$  & $0.077 $ &  $0.188 $ \\ [0.5ex]
\textsc{LWM} & $0.332$ & $0.312 $  & $0.322 $ &  $0.325 $  \\ [0.5ex]   
\hline
\textsc{ICICLE} & $\textbf{0.705}$ & $\textbf{0.753}$ & $\textbf{0.742}$ & $\textbf{0.728}$ \\[0.5ex]
\hline
\end{tabular}
\caption{Quantitative results for interpretability concept drift presented in \Cref{fig:fig1}. We compute IoU between similarities obtained after each task and after incremental tasks. E.g. in column ``task 1'', we calculate IoU between similarity maps of task one prototypes after each learning episode.}
\label{table:ious}
\vspace{-4ex}
\end{table}

\section{Related Works}

\paragraph{Continual Learning and Class Incremental Learning}

Existing continual learning methods can be broadly categorized into three types: replay-based, architecture-based, and regularization-based methods~\cite{delange2021continual, masana2022class}. Replay-based methods either save a small amount of data from previously seen tasks~\cite{bang2021rainbow, chaudhry2019tiny} or generate synthetic data with a generative model~\cite{wang2021ordisco, zhai2021hyper}. The replay data can be used during training together with the current data, such as in iCaRL~\cite{rebuffi2017icarl} and LUCIR~\cite{hou2019learning}, or to constrain the gradient direction while training, such as in AGEM~\cite{chaudhry2018efficient}. Architecture-based methods activate different subsets of network parameters for different tasks by allowing model parameters to grow linearly with the number of tasks. Previous works following this strategy include DER~\cite{yan2021dynamically}, Piggyback~\cite{mallya2018piggyback}, PackNet~\cite{mallya2018packnet}. Regularization-based methods add an additional regularization term derived from knowledge of previous tasks to the training loss. This can be done by either regularizing the weight space, which constrains important parameters~\cite{shi2021continual,tang2021layerwise}, or the functional space, which constrains predictions or intermediate features~\cite{douillard2020podnet, hu2021distilling}. EWC~\cite{kirkpatrick2017overcoming}, MAS~\cite{aljundi2018memory}, REWC~\cite{liu2018rotate}, SI~\cite{zenke2017continual}, and RWalk~\cite{chaudhry2018riemannian} constrain the importance of network parameters to prevent forgetting. Methods such as LWF~\cite{li2017learning}, LWM~\cite{dhar2019learning}, and BiC~\cite{wu2019large} leverage knowledge distillation to regularize features or predictions. Additionally, more challenging setups are considered in the field such as open-set interpretable continual learning~\cite{mundt2023wholistic}. When it comes to interpretable CL, the generative replay approaches~\cite{mundt2022unified} provide a certain degree of latent clarity. However, they require a decoder (for visualization) and may fail to produce realistic prototype images~\cite{chen2019looks}. Class-incremental learning (class-IL) is the most challenging scenario where the classifier learns new classes sequentialy. The model needs to maintain good performance on all classes seen so far~\cite{vandeven2019three}. Two types of evaluation methods are defined~\cite{masana2022class}: task-agnostic (no access to task-ID during inference, e.g., BiC~\cite{wu2019large}) and task-aware (task-ID is given during inference, e.g., HAT~\cite{serra2018overcoming}).

\begin{figure*}[t]
\centering
\includegraphics[width=0.9\linewidth]{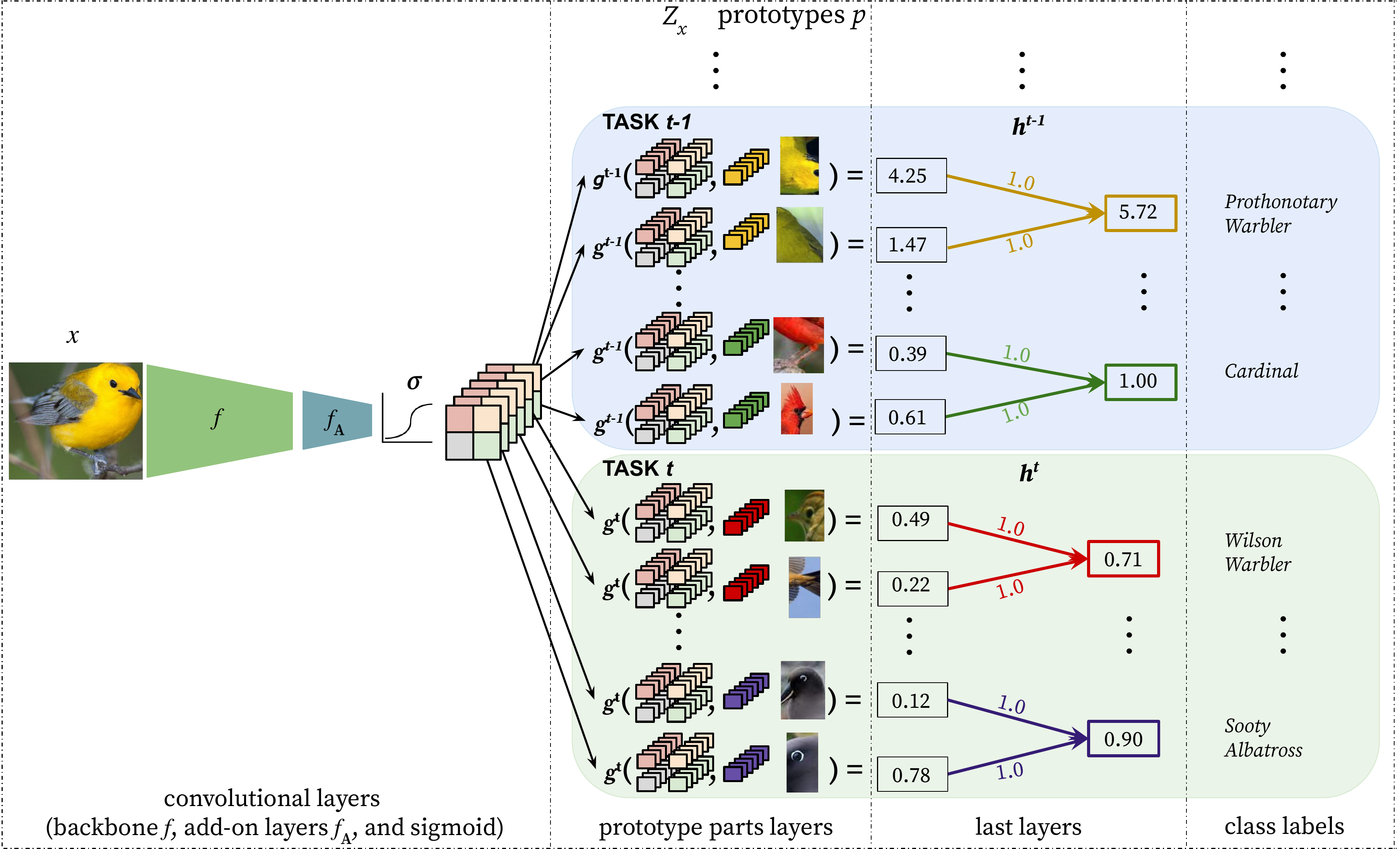}
\caption{The architecture of our ICICLE with separate prototypical part layers $g^t$ for each task $t$. In this example, prototypes of classes \textit{Prothonotary Warbler} and \textit{Cardinal} belong to task $t-1$, while prototypes of \textit{Wilson
Warbler} and \textit{Sooty Albatross} to task $t$. Layers $g^t$ are preceded by shared backbone $f$, add-on $f_A$, and sigmoid. Moreover, they are followed by the last layers $h^t$ with weight $h^t_{ci}=1$ if prototype $p_i$ is assigned to class $c$ and equals $0$ otherwise.
}
\label{fig:arch}
\vspace{-3ex}
\end{figure*}
\vspace{-2em}
\paragraph{Explainable Artificial Intelligence}
In the field of deep learning explanations, two types of models have been explored: post hoc and self-explainable models~\cite{rudin2019stop}. Post hoc models explain the reasoning process of black-box methods, including saliency maps~\cite{marcos2019semantically,rebuffi2020there,selvaraju2017grad,selvaraju2019taking,simonyan2014deep}, concept activation vectors~\cite{chen2020concept,NEURIPS2019_77d2afcb,kim2018interpretability,pmlr-v119-koh20a,NEURIPS2020_ecb287ff}, counterfactual examples~\cite{abbasnejad2020counterfactual,goyal2019counterfactual,mothilal2020explaining,niu2021counterfactual,wang2020scout}, and image perturbation analysis~\cite{basaj2021explaining,fong2019understanding,fong2017interpretable,ribeiro2016should}. Self-explainable models, on the other hand, aim to make the decision process more transparent and have attracted significant attention~\cite{NEURIPS2018_3e9f0fc9,brendel2018approximating}.
Recently, researchers have focused on enhancing the concept of prototypical parts introduced in ProtoPNet~\cite{chen2019looks} to represent the activation patterns of networks. Several extensions have been proposed, including TesNet~\cite{wang2021interpretable} and Deformable ProtoPNet~\cite{donnelly2022deformable}, which exploit orthogonality in prototype construction. ProtoPShare~\cite{rymarczyk2021protopshare}, ProtoTree~\cite{nauta2021neural}, and ProtoPool~\cite{rymarczyk2021interpretable} reduce the number of prototypes used in classification. Other methods consider hierarchical classification with prototypes~\cite{hase2019interpretable}, prototypical part transformation~\cite{li2018deep}, and knowledge distillation techniques from prototypes~\cite{keswani2022proto2proto}. Prototype-based solutions have been widely adopted in various applications such as medical imaging~\cite{afnan2021interpretable,barnett2021case,kim2021xprotonet,rymarczyk2021protomil,singh2021these}, time-series analysis~\cite{gee2019explaining}, graph classification~\cite{rymarczyk2022progrest,zhang2021protgnn}, sequence learning~\cite{ming2019interpretable}, and semantic segmentation~\cite{sacha2023protoseg}. In this work, we adapt the prototype mechanism to class incremental learning.

\vspace{-1em}
\section{Methods}
\vspace{-1em}
The aim of our approach is to increase the interpretability in the class-incremental scenario. For this purpose, we adapt prototypical parts~\cite{chen2019looks}, which directly participate in the model computation, making explanations faithful to the classification decision. 
To make this work self-contained, we first recall the prototypical parts methodology, and then we describe how we adapt them to the class-incremental scenario. We aim to compensate for interpretability concept drift, which we define at the end of this section. As we aim to compensate for interpretability concept drift, we define it at the end of this section.

\vspace{-1em}
\subsection{Prototypical parts methodology}
\paragraph{Architecture.}
The original implementation of prototypical parts~\cite{chen2019looks} introduces an additional prototypical part layer $g$ proceeded by a backbone convolutional network $f$ with an add-on $f_A$ and followed by the fully connected layer $h$. The $f_A$ add-on consists of two $1\times1$ convolutional layers and a sigmoid activation at the end, translating the convolutional output to a prototypical part space. The prototypical part layer $g$ consists of $K$ prototypes $p_i\in \mathbb{R}^D$ per class, and their assignment is handled by the fully connected layer $h$. If prototype $p_i$ is assigned to class $c$, then $h_{ci}=1$, otherwise, it is set to $-0.5$.
\vspace{-1em}
\paragraph{Inference.}
Given an input image $x$, the backbone $f$ generates its representation $f(x)$ of shape $H\times W\times D$, where $H$ and $W$ are the height and width of the representation obtained at the last convolutional layer, and $D$ is the number of channels. This representation is translated by $f_A$ to a prototypical part space, again of size $H\times W\times D$.
Then, each prototypical part $p_i$ is compared to each of $H\times W$ representation vectors to calculate the maximum similarity (i.e. the maximal activation of this prototype on the analyzed image) $\max_{j\in\{1..HW\}} sim(p_i, z_j)$, where $sim(p_i, z_j)=\log\frac{|z_j-p_i|_2 + 1}{|z_j-p_i|_2 + \eta}$ and $\eta \ll 1$. To obtain the final predictions, we push those values through the fully connected (and appropriately initialized) layer $h$.

\begin{figure}[tb]
\centering
\includegraphics[width=0.9\linewidth]{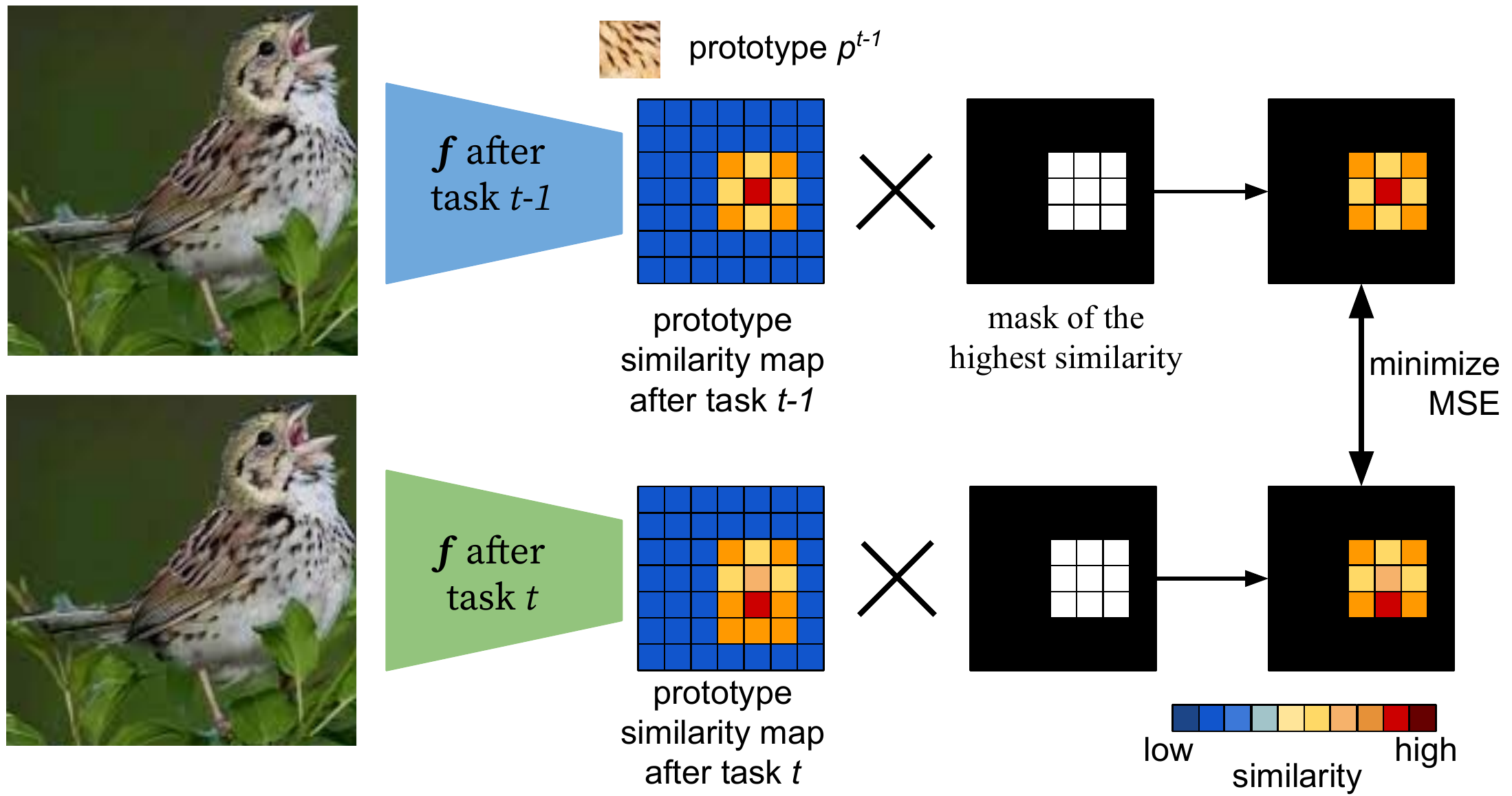}
\caption{Our interpretability regularization aims to minimize the changes in the prototype similarities. It takes a prototype $p^{t-1}$ of previous tasks and an image from task $t$, selects the image area with the highest similarity to this prototype (binary mask $S$), and punishes the model for any changes in this area caused by training task $t$.
}
\label{fig:distill}
\vspace{-4ex}
\end{figure}

\vspace{-3ex}
\paragraph{Training.} 
Training is divided into three optimization phases: warm-up, joint learning, and convex optimization of the last layer. The first phase trains add-on $f_A$ and the prototypical part layer $g$. The second phase learns $f_A$, $g$, and the backbone network $f$. The last phase fine-tunes the fully-connected layer $h$.
Training is conducted with the cross-entropy loss supported by two regularizations, cluster and separation costs~\cite{chen2019looks}. Cluster encourages each training image to have a latent patch close to at least one prototype of its class. In contrast, the separation cost encourages every latent patch of a training image to stay away from the prototypes of the remaining classes.

\vspace{-1ex}
\subsection{ICICLE}
\label{sec:icicle}
\vspace{-1ex}

Significant modifications of architecture and training are required to employ prototypical parts methodology to class-incremental learning (the inference is identical). Mostly because incremental learning has considerably different conjectures. It assumes $T$ tasks  $(C^1, X^1), (C^2, X^2), \ldots, (C^T, X^T)$, where each task $t$ contains classes $C^t$ and training set $X^t$. Moreover, during task $t$, only $X_t$ training data are available, as we consider the exemplar-free setup, where it is prohibited to save any data from previous tasks (no replay buffer is allowed).

\vspace{-2em}
\paragraph{Architecture.}
As in the baseline model, ICICLE comprises backbone $f$ and add-on $f_A$. However, it does not use one fixed prototypical part layer $g$ and one fully-connected layer $h$. Instead, it introduces a prototypical part layer $g^t$ and a fully-connected layer $h^t$ for each successive task. Layers $g^t$ consist of $M^t$ prototypical parts, where $M^t = K\cdot C^t$ and $K$ is the number of prototypes per class. On the other hand, layer $h^t$ has weight $h^t_{ci}=1$ if prototype $p_i$ is assigned to class $c$ and it is set to $0$ otherwise. We eliminated negative weights ($-0.5$) from the last layer because multi-stage training is not beneficial for a class-incremental scenario (see \Cref{fig:ll}).

\vspace{-3ex}
\paragraph{Training.}
To prevent catastrophic forgetting, ICICLE modifies the loss function of the baseline solution. Additionally, it introduces three mechanisms: interpretability regularization, proximity-based prototype initialization, and task-recency bias compensation.
Regarding the baseline loss function, the cross-entropy is calculated on the full output of the model, including logits from classes learned in previous tasks. However, the cluster and separation costs are only calculated within the $g^t$ head.

\begin{figure}[t]
\centering
\includegraphics[width=0.85\linewidth]{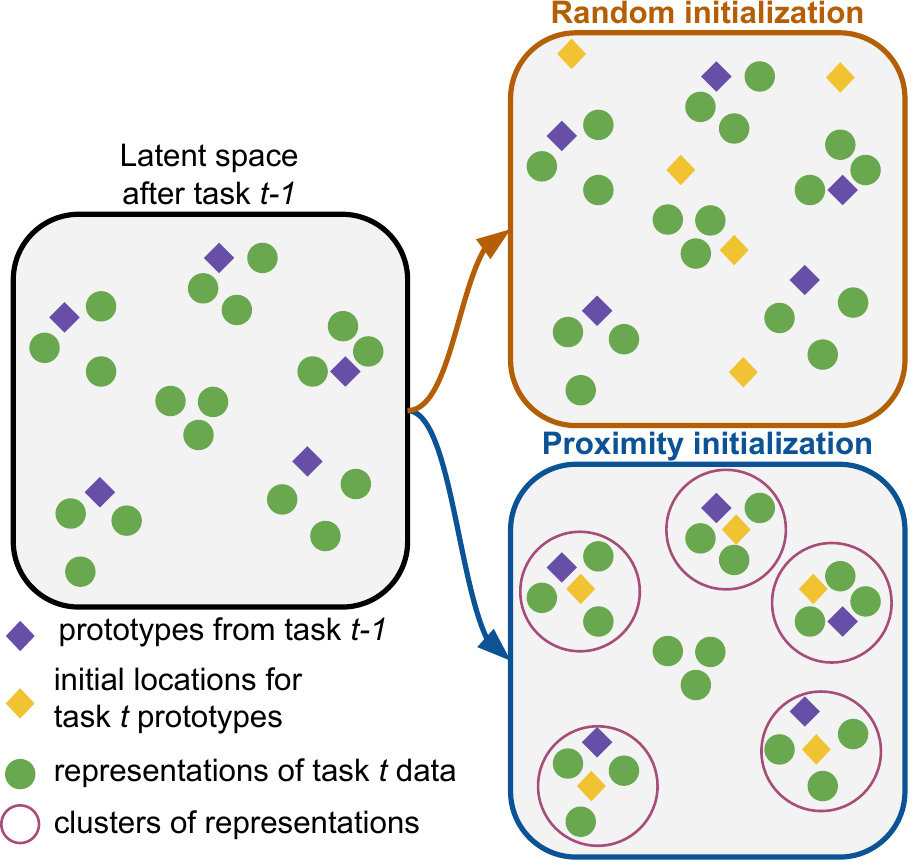}
\caption{We introduce a new proximity-based prototype initialization. It starts by passing training samples of task $t$ through the network (green dots) and choosing representations closest to existing prototypes (violet diamonds). This results in many points, which we cluster (purple circles) to obtain the initial locations of task $t$ prototypical parts (yellow diamonds). Such initialization (bottom right) is preferred over random initialization (top right), where new prototypes can be created far from the old ones, even though they are only slightly different.
}
\label{fig:proximity}
\vspace{-4ex}
\end{figure}

\noindent \emph{Interpretability regularization.}
Knowledge distillation~\cite{hinton2014distilling} is one of the strong regularization methods applied to prevent forgetting~\cite{li2017learning}. However, the results obtained by its straightforward application are not satisfactory and lead to significant interpretation drift (see \Cref{fig:fig1} and \Cref{section:results}). Therefore, we introduce an additional regularization cost $L_{IR}$ (see ~\Cref{fig:distill}), inspired by~\cite{keswani2022proto2proto}, that minimizes the changes in the similarities for the prototypical parts of the previous tasks. It is defined as:
\vspace{-1em}
\begin{equation}
L_{IR} = \sum_{i=0}^{H}\sum_{j=0}^{W}|sim(p^{t-1}, z_{i,j}^{t}) - sim(p^{t}, z_{i,j}^t)| \cdot S_{i,j}
\vspace{-1em}
\end{equation}
where $sim(p^{t-1}, z_{i,j}^{t})$ is computed for the model stored before training task $t$, and $S$ is a binary mask of size $H\times W$, indicating the representation pixels with the highest similarity ($\gamma$ quantile of those pixels). Such similarity distillation gives higher plasticity when learning a new task but, at the same time, reduces the interpretability drift.

\noindent \emph{Proximity-based prototype initialization.}
Random initialization of prototypes, proposed in~\cite{chen2019looks}, fails in the incremental learning (see \Cref{table:init_strategies}). Most probably because new prototypes can be created far from the old ones, which are only slightly different (e.g. wing prototypes of various bird species). Therefore, we introduce initialization that sets new prototypes close to the old ones (see \Cref{fig:proximity}). We start by passing training samples of task $t$ through the network and determining which patches are the most similar to existing prototypes (we choose patches from the $\alpha$ quantile). More specifically, we compute set $\{z_j^t: \max_{i} sim(p_i^{t-1}, z_j^t) \in \alpha\text{ quantile}\}$. This results in many candidates for new task prototypical parts. To obtain $K\cdot C^t$ prototypes, we perform KMeans++ clustering, and the resulting centers of clusters are used to initialize the prototypical parts in $g^t$.

\noindent \emph{Task-recency bias compensation.}
When the model learns task $t$, the similarities to the prototypes of previous tasks drop, mostly due to the changes in the backbone (see \Cref{fig:compensate}). That is why, after training the final task, we compensate for this using $T-1$ constants obtained using the last tasks data. {More precisely, for each of the previous tasks $t<T$, we take logits $y^t=h^t \circ g^t \circ f(x)$ obtained for all $x\in X^T$ and calculate bias $c^t$ so that $|\{x \in X^T: \max{(y^t+c^t)} > \max{y^T}\}| = u|X^T|$. Intuitively, we adjust $c^t$ so the model changes $u\%$ of its prediction from task T to task t. We determined experimentally that $u=10\%$ is optimal.

\begin{table*}[t]

\centering
\scriptsize 
\begin{tabular}{|c||c|c|c|c|c|c|}
\hline
& \multicolumn{3}{c|}{\textsc{Avg. inc. task-aware accuracy}} & \multicolumn{3}{c|}{\textsc{Avg. inc. task-agnostic accuracy}} \\ [0.5ex]
\cline{2-7}
\textsc{Method} & \textsc{$4$ tasks} & \textsc{$10$ tasks} & \textsc{$20$ tasks} & \textsc{$4$ tasks} & \textsc{$10$ tasks} & \textsc{$20$ tasks}  \\ [0.5ex]
\hline
\hline
\textsc{Freezing} & $0.560 \pm 0.027$
& $0.531 \pm 0.042$ & $0.452 \pm 0.055$ & $0.309 \pm 0.024$ & $0.115 \pm0.028$ & $0.078 \pm 0.004$ \\ [0.5ex]
\textsc{Finetuning} & $0.229 \pm 0.005$  & $0.129 \pm 0.017$ & $0.147 \pm 0.021$ & $0.177 \pm 0.006$ & $0.072 \pm 0.008$ & $0.044 \pm 0.006$ \\ [0.5ex] 
\textsc{EWC} & $0.445\pm 0.012$ & $0.288 \pm 0.034$ & $0.188 \pm 0.031$
& $0.213\pm 0.008$ & $0.095 \pm 0.007$ & $0.046 \pm 0.011$ \\ [0.5ex]
\textsc{LWM} & $0.452 \pm 0.023$ & 
$0.294 \pm 0.032$ & $0.226 \pm 0.025$
& $0.180 \pm 0.028$ & $0.090 \pm 0.011$ & $0.044 \pm 0.008$\\ [0.5ex]
\textsc{LWF} & $0.301\pm0.048$ & $0.175 \pm 0.028$ & $0.129 \pm 0.023$
&  $0.219 \pm 0.019$ & $0.078 \pm 0.008$ & $0.072 \pm 0.008$ \\ [0.5ex]
\hline
\textsc{ICICLE} & $\textbf{0.654} \pm \textbf{0.011}$ & $\textbf{0.602} \pm \textbf{0.035}$ & $\textbf{0.497} \pm \textbf{0.099}$
& $\textbf{0.350} \pm \textbf{0.053}$ & $\textbf{0.185} \pm \textbf{0.005}$ & $\textbf{0.099} \pm \textbf{0.003}$ \\[0.5ex]
\hline
Multi-task & $0.858 \pm 0.005$ & $0.905\pm0.012$ & $0.935 \pm 0.019$ & $0.499 \pm 0.009$ & $0.196 \pm 0.017$ & $0.148 \pm 0.009$ \\ [0.5ex]
\hline
FeTrIL~\cite{petit2023fetril} & $0.750 \pm 0.008$ & $0.607 \pm 0.018$ & $0.407 \pm 0.051$ & $0.375 \pm 0.006$ & $0.199 \pm 0.003$ & $0.127 \pm 0.011$  \\ [0.5ex]
PASS ~\cite{zhu2021prototype} & $\underline{0.775 \pm 0.006}$ & $\underline{0.647 \pm 0.003}$ & $\underline{0.518 \pm 0.012}$ & $\underline{0.395 \pm 0.001}$ & $\underline{0.233 \pm 0.009}$ & $\underline{0.139 \pm 0.017}$ \\ [0.5ex]
         \hline
\hline
\end{tabular}

\caption{Average incremental accuracy comparison for different numbers of tasks on CUB-200-2011, demonstrating the negative impact of the high number of tasks to be learned on models' performance. Despite this trend, ICICLE outperforms the baseline methods across all task numbers. Additionally, we show the gap between interpretable and black-box models by comparing ICICLE to FeTrIL and PASS.}
\label{table:cub_tasks}
\vspace{-4ex}
\end{table*}
\subsection{Interpretability Concept Drift}
As noted in the caption of Figure~\ref{fig:fig1}, the interpretability concept drift occurs when a similarity map differs between tasks. Therefore, it can be formally defined as: 
\vspace{-0.5em}
\begin{equation}
ICD = \mathbb{E}_{i,j=1}^{H,W} \left|sim(p^{t-1}, z_{i,j}^{t}) - sim(p^{t}, z_{i,j}^t)\right|,
\vspace{-0.5em}
\end{equation}
where $(z_{i,j})_{i,j=1}^{H,W}$ corresponds to input image representation, $p^{t-1}$ and $p^t$ correspond to prototypical part $p$ before and after task $t$, and $sim$ is similarity defined in Section~3.1 of the paper. Thus, the greater $ICD$, the greater the interpretability concept drift.
\vspace{-1em}
\section{Experimental Setup}
We evaluate our approach on the CUB-200-2011~\cite{wah2011caltech} and Stanford Cars~\cite{krause20133d} datasets to classify 200 bird species and 196 car models, respectively. We consider 4, 10, and 20 task learning scenarios for birds and 4, 7, and 14 options for cars. As the backbone $f$, we take ResNet-34~\cite{he2016deep} without the last layer and pre-trained on ImageNet~\cite{deng2009imagenet}. We set the number of prototypes per class to $10$. Moreover, we use prototypical parts of size $1 \times 1 \times 256$ and $1 \times 1 \times 128$ for birds and cars, respectively. The weights of CE, cluster, separation, and distillation costs in the loss function equal $1.0$, $0.8$, $-0.08$, and $0.01$. In distillation, we take $\lambda=1/49$ representation pixels with the highest similarity. For proximity-based initialization, we use $\alpha = 0.5$. For task-recency bias compensation, we take $c_t$, which changes the predictions of the last validation set by less than $10\%$.  
As the implementation framework, we use FACIL~\cite{masana2022class} based on the PyTorch library\footnote{\url{https://pytorch.org}}. Details on the experimental setup are provided in the Supplementary Materials\footnote{Code available at: \url{https://github.com/gmum/ICICLE}}.

\begin{figure}[t]
\centering
\includegraphics[width=0.99\linewidth]{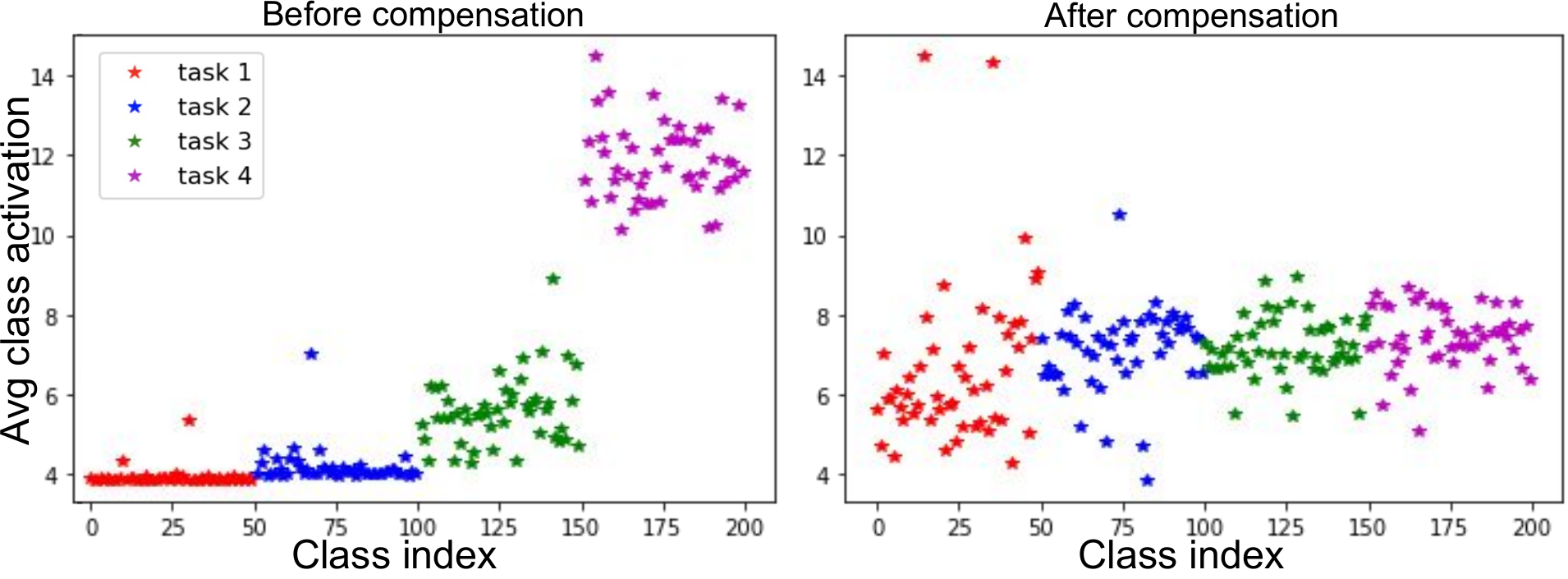}
\caption{When the model learns task $t$, the similarities to the prototypes of previous tasks drop and are significantly lower than those of new tasks (upper plot). That is why, after training the final task, we compensate it with $T-1$ calculated constants. As a result, the similarities obtained by prototypes of all tasks are roughly equalized.
}
\label{fig:compensate}
\vspace{-4ex}
\end{figure}

\begin{table*}[t]

\centering
\scriptsize 
\begin{tabular}{|c||c|c|c|c|c|c|c|c|}
\hline
& \multicolumn{4}{c|}{\textsc{Task-aware Accuracy}} & \multicolumn{4}{c|}{\textsc{Task-agnostic Accuracy}} \\ [0.5ex]
\cline{2-9}
\textsc{Method} & \textsc{Task 1 } & \textsc{Task 2 } & \textsc{Task 3 } & \textsc{Task 4 } & \textsc{Task 1 } & \textsc{Task 2 } & \textsc{Task 3 } & \textsc{Task 4 }  \\ [0.5ex]
\hline
\hline
{\textsc{Freezing}} &  $\textbf{0.806}\pm \textbf{0.024}$ & $0.462\pm0.037$ & $0.517 \pm 0.041$ & $0.455 \pm 0.027$ &  $\textbf{0.570}\pm \textbf{0.031}$ & $0.195\pm 0.017$ & $0.258\pm 0.019$ & $0.213\pm 0.020$  \\ [0.5ex]
{\textsc{Finetuning}} & $0.007\pm0.004$ & $0.016\pm0.008$ & $0.032\pm 0.009$ & $0.759\pm0.019$ & $0.0\pm 0.0$ & $0.0\pm 0.0$ & $0.0\pm 0.0$ & $0.759\pm 0.019$   \\ [0.5ex]
\textsc{EWC} & $0.244 \pm 0.024$ & $0.378\pm 0.072$ & $0.539\pm 0.043$ & $0.602\pm 0.054$
& $0.001\pm 0.001$ & $0.059\pm 0.004$ & $0.267\pm 0.031$ & $0.527\pm 0.051$ \\ [0.5ex]
\textsc{LWF} & $0.169 \pm 0.046$ & $0.119 \pm 0.008$ & $0.235\pm 0.017$ & $0.743 \pm 0.061$ & $0.158\pm0.035$ & $0.003 \pm 0.002$ & $0.018\pm 0.003$ & $0.537 \pm 0.142$  \\ [0.5ex]
\textsc{LWM} & $0.195 \pm 0.012$ & $0.412 \pm 0.014$ & $0.430 \pm 0.028$ & $\textbf{0.772} \pm \textbf{0.011}$ 
& $0.027\pm 0.024$ & $0.023\pm 0.020$ & $0.085 \pm 0.005$ & $\textbf{0.772} \pm \textbf{0.037}$ \\ [0.5ex]
\hline
\textsc{ICICLE} & $0.523 \pm 0.020$ & $\textbf{0.663} \pm \textbf{0.053}$ & $\textbf{0.709} \pm \textbf{0.038}$ & $0.723 \pm 0.002$ & $0.233 \pm 0.014$ & $\textbf{0.365} \pm \textbf{0.021}$ & $\textbf{0.314} \pm \textbf{0.011}$ & $0.486 \pm 0.021$ \\ [0.5ex]
\hline

\end{tabular}
\caption{Comparison of task accuracies for modified ProtoPNet architecture in a class-incremental learning scenario after 4 tasks train on CUB-200-2011 dataset, averaged over 3 runs with standard error of the mean. Our ICICLE outperforms baseline methods and achieves the best results for all previous incremental tasks, demonstrating its ability to maintain prior knowledge while learning new tasks. Freezing due to the weight fixation cannot properly learn new tasks.}
\label{table:cub_results4}
\vspace{-5ex}
\end{table*}
\vspace{-1em}
\section{Results}
\label{section:results}

\paragraph{Performance.} We evaluated the effectiveness of ICICLE by comparing it with commonly used exemplar-free baseline methods in class-incremental learning, including LWF~\cite{li2017learning}, LWM~\cite{dhar2019learning}, and EWC\cite{kirkpatrick2017overcoming}\footnote{Extending interpretable models with more complicated exemplar-free methods is not straightforward, and we, therefore, excluded methods such as SDC~\cite{yu2020semantic} (which requires learning with a metric loss) and PASS~\cite{zhu2021prototype} (which requires a combination with self-supervised learning).}. Additionally, Finetuning, and Freezing of the feature extractor (not trained at all) are provided. We also report multitask learning as an upper-bound where the various tasks are learned jointly in a multitask manner. To do so, we analyzed task-aware and task-agnostic accuracy for each task after the last one (\Cref{table:cub_results4}) and the aggregated incremental average accuracies after learning the last task in scenarios involving 4, 10, and 20 tasks for CUB (\Cref{table:cub_tasks}) and 4, 7, and 14 tasks for Stanford Cars (Supplementary Materials). All methods use the same feature extractor network architectures and ProtoPNet for prototypical part-based learning. Our method outperformed the baseline methods in all cases, indicating its superior performance for prototypical part-based learning in a continual manner. ICICLE retains knowledge from previous tasks better, which results in a more balanced accuracy between tasks and higher accuracy for the first task compared to all other approaches. However, despite the significant improvement, our approach still has room for improvement compared to the upper-bound of multi-task training. With a larger number of tasks, the forgetting of the model is higher, resulting in poorer results, which may be attributed to the number of details that prototypes need to capture to classify a task correctly. Furthermore, we have noticed that freezing is a robust baseline for a task-aware scenario because of the model's fixed nature and pretrained backbone.

\vspace{-1em}
\paragraph{Interpretability} To evaluate if the prototype's graphical representation of the concept has changed and how much we use the IoU metric~\cite{sacha2023protoseg}. IoU measures what is the overlap of the prototype visual representation (like in ~\Cref{fig:fig1}) from the task in which it was learned, through all the following tasks. Freezing is superior in preserving the prototypical information because all weights from previous tasks are fixed. In terms of methods allowing changes in backbone and previously learned prototypes, ICICLE is superior over all baselines, as shown in \Cref{table:ious}. ICICLE keeps interpretable prototypes consistent with interpretability regularization distilling already learned concepts.



\subsection{Ablation study and analysis}

\paragraph{Why changes in ProtoPNet architecture and training are needed?} 

ProtoPNet in the last training stage (last layer convex optimizations) aims to finetune positive connections and regularize the negative to be 0. As a result, the converged model returns interpretations in the form of positive reasoning, desired by the end users~\cite{rudin2022interpretable}. In the CL setting, the last step of training changes the negative connections in a different manner (see ~\Cref{fig:ll}). On the other hand, in an exemplar-free continual learning scenario, conducting the last-layer learning phase is unfeasible at the end of the training. That is why, we modified the ProtoPNet's last layer and retained only positive connections initialized to 1, eliminating the need for the convex optimization step.

\begin{table}[t]

\centering
\scriptsize 
\begin{tabular}{|c|c|c||c|c|}
\hline
Regularization & Initialization & Compensation & TAw acc. & TAg acc. \\ [0.5ex]
\hline
 &   &  & 0.216 & 0.182 \\ 
\checkmark &   &  & 0.559 & 0.280 \\ 
\checkmark &  \checkmark &  & \textbf{0.654} & 0.335 \\ 
\checkmark &  \checkmark & \checkmark & \textbf{0.654} & \textbf{0.350} \\ [0.5ex]

\hline
\end{tabular}

\caption{Influence of different novel components on the average incremental accuracy in four tasks learning scenario. Combination of all components results in the best-performing model.}
\label{table:novelties_infuence}
\vspace{-1ex}
\end{table}

\begin{figure}[t]
\centering
\includegraphics[width=0.9\linewidth]{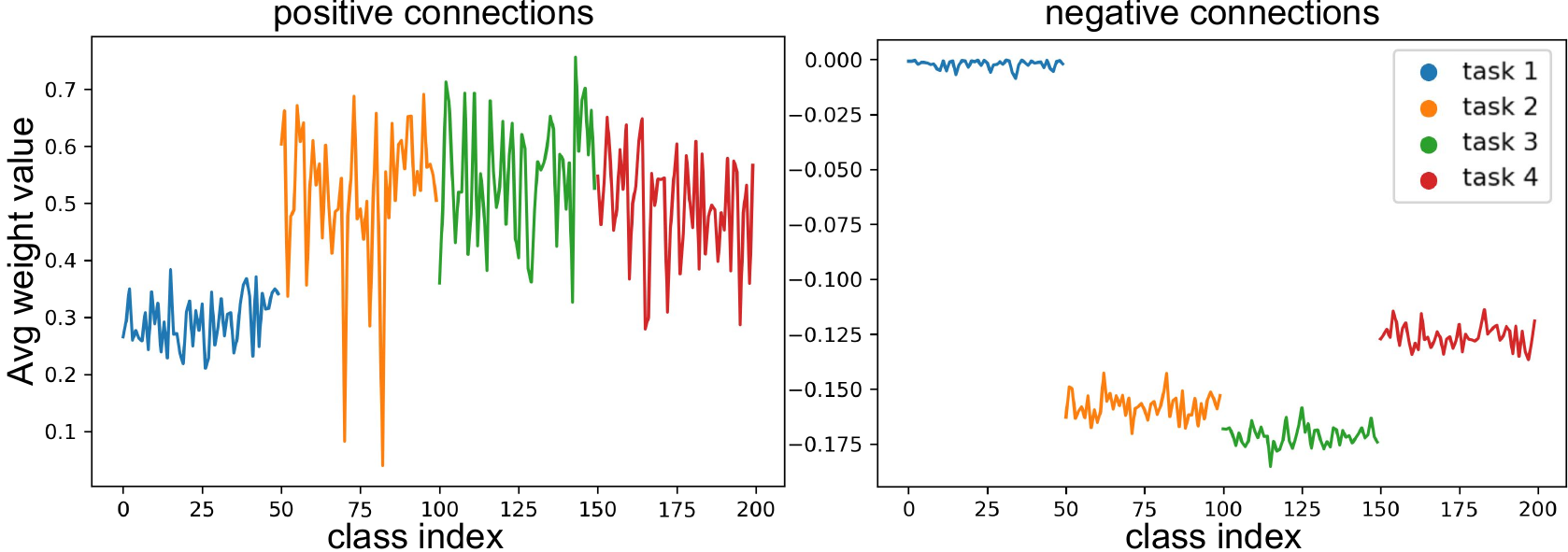}
\caption{Average weight of positive and negative connections per class in 4 task learning scenario. Unbalanced and strong negative connections between tasks result in undesired properties in terms of the model's interpretability.}
\label{fig:ll}
\vspace{-3ex}
\end{figure}
\vspace{-1em}
\paragraph{What is the influence of each of the introduced components?}
~\Cref{table:novelties_infuence} presents the influence of different components of our approach on the final model's average incremental accuracy in the CUB-200-2011 dataset with four tasks split scenario. Combining all the components resulted in the best-performing model. Our results show that compensation of task-recency bias helps in task-agnostic evaluation and gives additional improvement of $4.5\%$. However, most of the accuracy improvements were attributed to interpretability regularization and proximity initialization. Notably, task-recency bias compensation significantly improved the performance of task one classes compared to an approach without it, from 0.028 to 0.255 in a task-agnostic scenario, as detailed in the Supplementary Materials.

\begin{figure}[t]
\centering
\includegraphics[width=0.9\linewidth]{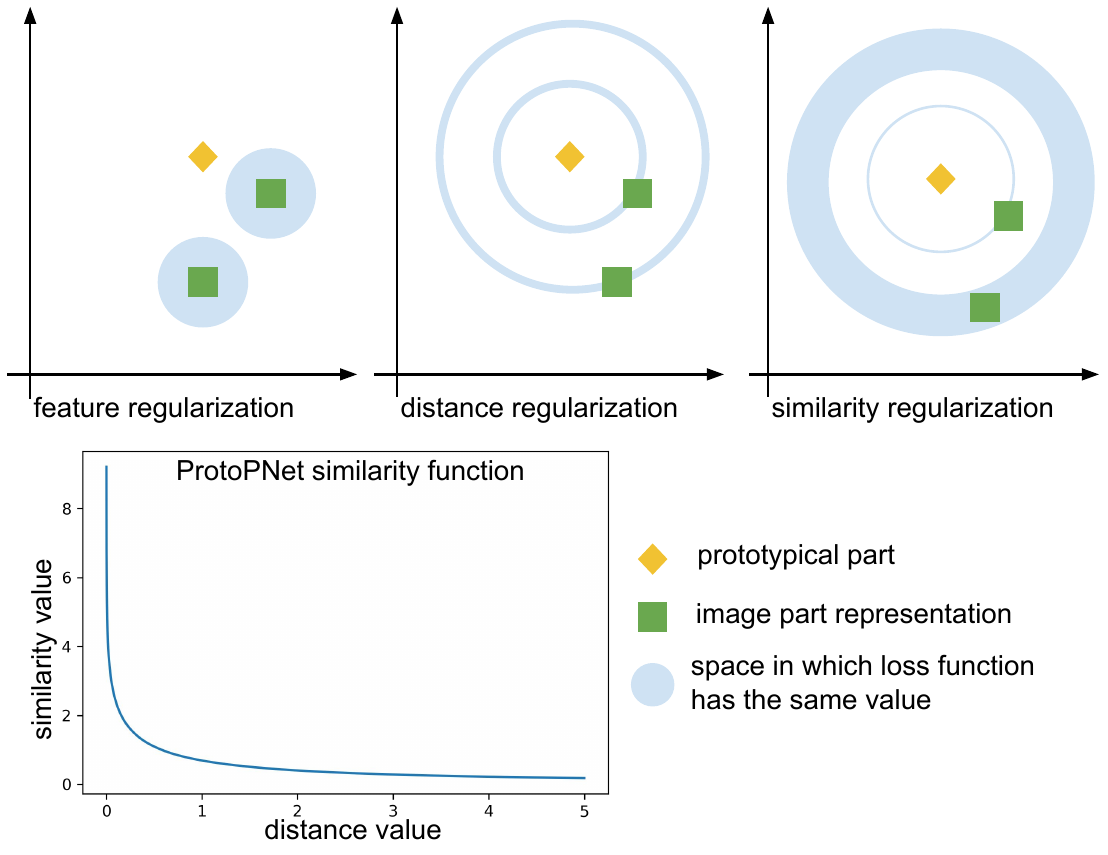}
\caption{Visualization of three possible approaches to interpretability similarity and their influence on the model plasticity. Only similarity-based regularization takes into account how a given image part corresponds to a prototypical part. If it is close then the similarity value is high and small changes in the distance results in a great decrease in similarity. While latent vectors that are distant from prototypical parts can more freely be changed by the model to better represents the current task data. Other approaches are limiting the models' plasticity treating each latent representation of the image part as equally important.}
\vspace{-2em}
\label{fig:reg_place}

\end{figure}
\vspace{-2em}
\paragraph{Where should we perform interpretability regularization?} 
The ProtoPNet model's prototypical layer can be regularized in three different ways: feature regularization on add-on layer representations, regularization of distances between prototypical parts and latent data vectors, and similarity-based regularization. The strictest approach is feature regularization, which does not allow the model to change how it represents data from a new task, resulting in significantly reduced model plasticity. When distances are regularized, the model can change its representation to maintain the same distance from the prototype on the surface of the sphere. On the other hand, similarity-based regularization allows the model to retain key knowledge from previous tasks by preserving only the information related to specific features that are close to the prototypical parts in the latent space, allowing for greater flexibility in exchange for forgetting irrelevant features. Therefore, we stick to interpretability regularization in ICICLE, which is based on similarities and maintains the essential knowledge from previous tasks while retaining high plasticity to learn new ones. \Cref{fig:reg_place} illustrates these three approaches and their comparison in terms of average incremental accuracy for ProtoPNet only with regularization (without changing initialization): $0.507$, $0.535$, and $0.559$ in task-aware and $0.261$, $0.230$, and $0.280$ in task-agnostic scenarios for feature, distance, and similarity-based regularization, respectively, on the CUB-200-2011 dataset with four tasks scenarios.

\vspace{-2em}
\paragraph{What is the influence of hyperparameters in interpretability regularization?} 
In \Cref{fig:lambda} and \Cref{fig:mask_size} the influence of $\lambda$ and mask percentile threshold in the interpretability regularization on average incremental accuracies are presented. We use CUB-200-2011 datasets with four tasks split setting. 
For this dataset, the results reveal that the regularization of only the maximum prototypical similarity is the most effective (\Cref{fig:mask_size}). Regarding $\lambda_{IR}$, a value that is too small leads to high network plasticity, increased forgetting, and poor results, while a value that is too large reduces model plasticity and may not represent new knowledge well.

\vspace{-2em}
\paragraph{Which way is the best to initialize new prototypical parts?} In this ablation part, we investigate the optimal strategy for initializing prototypical parts at the beginning of a new task in the ProtoPNet model. We evaluate our initialization method, which initializes the parts in close proximity to existing prototypes, against three other approaches: random initialization, clustering of all image part representations, and clustering of only distant latent vectors. Results are presented in \Cref{table:init_strategies}. The proximity initialization method outperforms the distant strategy, as the latter tends to assign prototypical parts to latent vectors that correspond to the background of the images, resulting in learning irrelevant concepts that can easily activate on other task data, as shown in the Supplementary Materials.

\begin{table}[t]

\centering
\scriptsize 
\begin{tabular}{|c||c|c|c|c|}
\hline
Initialization type & Random & Distant & All & Proximity \\ [0.5ex]
\hline
 Task aware acc. & $0.559$  & $0.592$ & $0.626$ & $\textbf{0.654}$ \\ [0.5ex]
 Task agnostic acc. & $0.280$  & $0.290$ & $0.297$ & $\textbf{0.335}$ \\ [0.5ex]
\hline
\end{tabular}
\caption{Comparison of different initialization strategies for prototypical parts. Our proximity initialization of new task prototypes is superior.}
\label{table:init_strategies}
\vspace{-2em}
\end{table}

\vspace{-2em}
\paragraph{Does ICICLE generalize to other architectures?} Lastly, we show that ICICLE generalizes to other concept-based architecture. We demonstrate that using a TesNet model~\cite{wang2021interpretable}, and provide results in \Cref{table:tesnet}, where ICICLE obtains the best results. The average incremental accuracy of ICICLE with TesNet is even better than ProtoPNet for both task-aware and task-agnostic evaluation.

\vspace{-1em}
\section{Conclusions and future work}

This work proposes a novel approach called ICICLE for interpretable class incremental learning. ICICLE is based on prototypical parts and incorporates interpretability regularization, proximity initialization, and compensation for task-recency bias. The proposed method outperforms classical class-incremental learning methods applied for prototypical part-based networks in terms of task-aware and task-agnostic accuracies while maintaining prototype interpretability. We also conducted ablation studies and multiple analyses to justify our choices and highlight the challenges associated with combining interpretable concepts with CL. This work is expected to inspire research on XAI and CL.

\begin{table}[t]

\centering
\scriptsize 
\begin{tabular}{|c||c|c|c|c|c|c|}
\hline
 & Freezing & Finetuning & EWC & LWM & LWF & ICICLE \\ [0.5ex]
\hline
TAw Acc. & $0.637$ & $0.355$ & $0.592$ & $0.648$ & $0.581$ & $\textbf{0.746}$ \\ [0.5ex]
TAg Acc. & $0.222$ & $0.183$ & $0.272$ & $0.252$ & $0.205$ & $\textbf{0.362}$ \\ [0.5ex]
\hline
\end{tabular}
\caption{Results for four task learning scenario on CUB-200-2011 dataset with TesNet~\cite{wang2021interpretable} as a concept-based architecture. The table shows the versatility of the ICICLE approach for interpretable models.}
\label{table:tesnet}
\vspace{-1em}
\end{table}

\begin{figure}[t]
\centering
\includegraphics[width=0.9\linewidth]{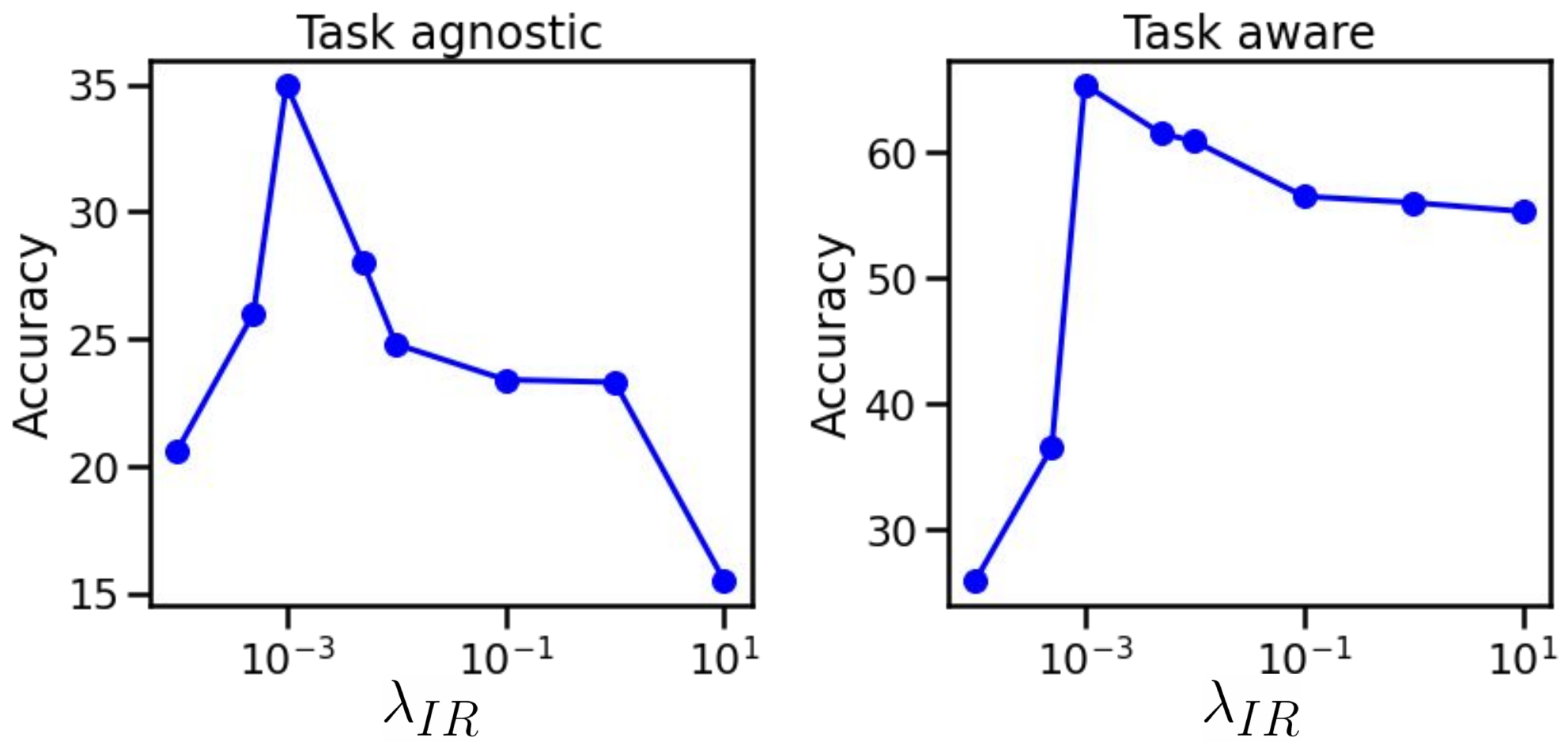}
\caption{Influence of the $\lambda_{IR}$ in the interpretability regularization.}
\label{fig:lambda}
\vspace{-2em}
\end{figure}

\begin{figure}[t]
\centering
\includegraphics[width=0.9\linewidth]{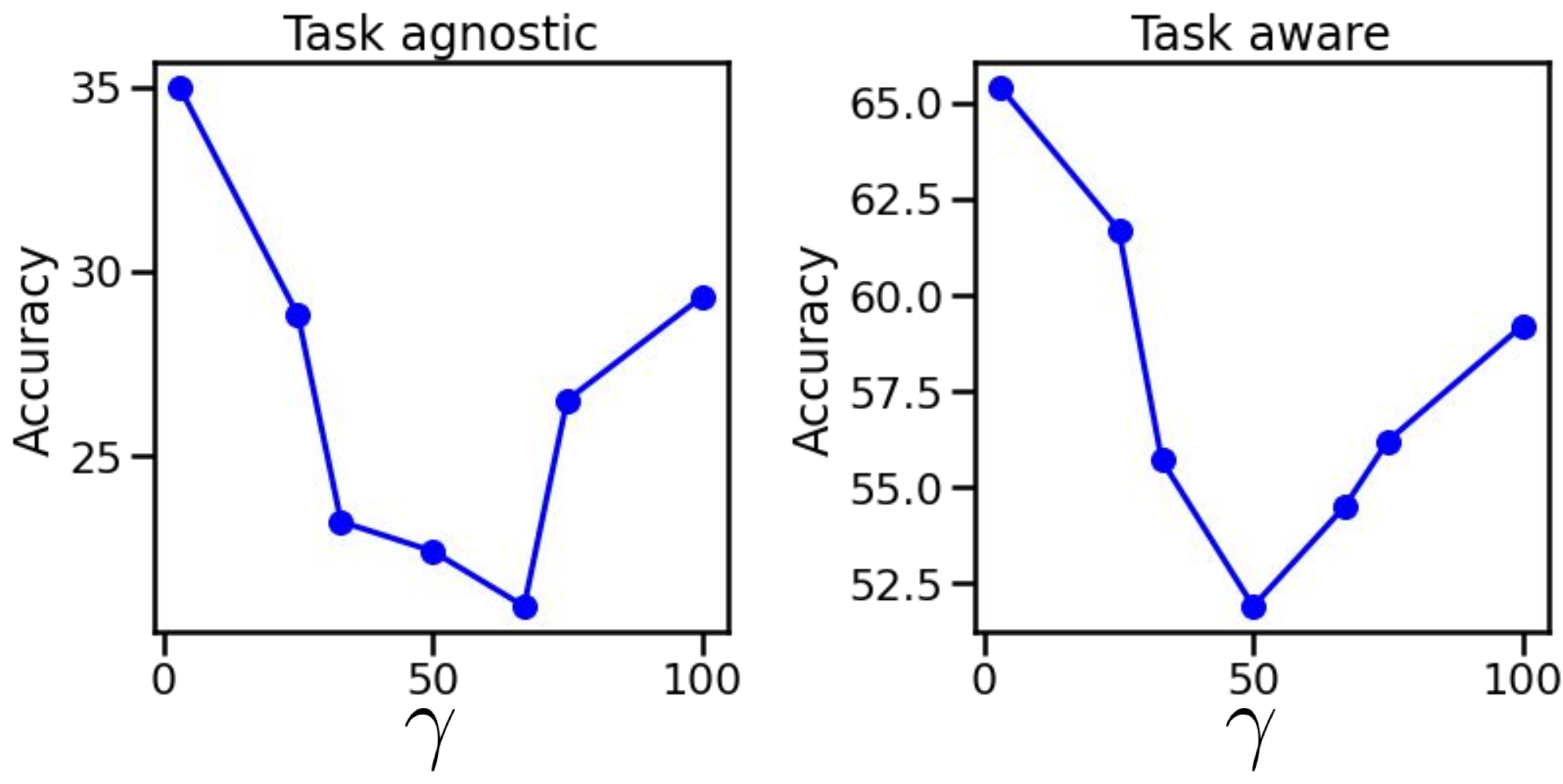}
\caption{Influence of $\gamma$ in the interpretability regularization. Notice that regularizing only in the place of maximum similarity is the most beneficial for ICICLE for the four task learning scenario in CUB-200-2011.}
\label{fig:mask_size}
\vspace{-1em}
\end{figure}

Moving forward, we plan to explore methods suitable for single-class incremental learning with interpretable models. We also intend to investigate how other interpretable architectures, such as B-COS~\cite{bohle2022b}, can be adapted to the class incremental learning scenario.

\vspace{-1em}
\paragraph{Limitations.}
Our work is limited to prototypical part methods, that are suited for fine-grained image recognition and inherits their drawbacks previously discussed in~\cite{gautam2023looks,hoffmann2021looks,kim2022hive,nauta2022looks,rymarczyk2021interpretable}.However, recently there are first works generalizing them to standard datasets (not fine-grained)~\cite{nauta2023pip}. Additionally, as we consider only an exemplar-free scenario and closed-set recognition, we do not analyze how having a replay buffer would influence the method's performance and how this method would fit in the open-set settings.

\vspace{-1em}
\paragraph{Impact.}
ICICLE highlights that traditional exemplar-free approaches for continual learning are not well suited for gray-box models that utilize concepts for predictions. This finding has implications for the development of continual learning methods, as they must balance the need for generality with the need to be adapted to specific architectures. Furthermore, it has an impact on the field of concept-based models and explainable AI, demonstrating the need for further research on CL methods for XAI. In some cases, practitioners who know that their system will need to learn new tasks continuously may choose to use black-box models and explainers rather than interpretable models, sacrificing the fidelity of explanations for improved model performance.

\vspace{-1em}
\section*{Acknowledgements}
Joost van de Weijer and Bartłomiej Twardowski acknowledge the support from the Spanish Government funding for projects PID2019-104174GB-I00, TED2021-132513B-I00, and grant RYC2021-032765-I, the European Commission under the Horizon 2020 Programme, funded by MCIN/AEI/10.13039/501100011033 and by European Union NextGenerationEU/PRTR.
The work of 
Dawid Rymarczyk was supported by the National Science Centre (Poland), grant no. 2022/45/N/ST6/04147. He also received an incentive scholarship from the program Excellence Initiative - Research University funds at the Jagiellonian University in Kraków.
The work of Bartosz Zieliński was funded by the National Science Centre (Poland) grant no. 2022/47/B/ST6/03397 and the research project “Bio-inspired artificial neural network” (grant no. POIR.04.04.00-00-14DE/18-00) within the Team-Net program of the Foundation for Polish Science co-financed by the European Union under the European Regional Development Fund.
Finally, some experiments were performed on servers purchased with funds from a Priority Research Area (Artificial Intelligence Computing Center Core Facility) grant under the Strategic Programme Excellence Initiative at Jagiellonian University.

{\small
\bibliographystyle{ieee_fullname}
\bibliography{egbib}
}
\newpage
\section*{Additional experimental setup details}

Here we present additional details on the experimental setup. We performed a hyperparameter search for $\lambda_{dist}$ ($\lambda_{dist} \in \{10.0, 5.0, 0.0, 0.5, 0.1, 0.05, 0.01, 0.005, 0.001, 0.0005, \\ 0.0001\}$). We use Adam optimizer with a learning rate of $0.001$ and parameters $\beta_1=0.9$ and $\beta_2=0.999$. We set the batch size to $75$ and use input images of resolution $224\times224\times3$. The weights of the network are initialized with Xavier's normal initializer.

We perform a warmup training where the weights of $f$ are frozen for $10$ epochs, and then we train the model until it converges with $12$ epochs early stopping. We use the learning schema presented in Table~\ref{tab:learning_scheme}. Depending on the number of tasks, we perform warm-up training with $\{5, 5, 4\}$ epochs and joint training phase for $\{21, 15, 11\}$, longer with fewer tasks. Similarly, we perform prototype projection every $\{10, 7, 5\}$ epoch. So with more tasks, we perform fewer training epochs (Table~\ref{tab:learning_scheme}). 

\begin{table*}[]
    \centering
    \fontsize{8}{10}\selectfont
    \begin{tabular}{l@{\;}ccccc}
    \hline
         Phase & Model layers & Learning rate & Scheduler & Weight decay & Duration \\
         \hline
         \multirow{2}{*}{Warm-up} & add-on $1\!\!\times\!\!1$ convolution & $1\cdot 10^{-3}$ & \multirow{2}{*}{None} & \multirow{2}{*}{None} & \multirow{2}{*}{$5, 5, 4$ epochs}\\
         & prototypical layer & $1\cdot 10^{-3}$ \\
         \hline
         \multirow{3}{*}{Joint} & convolutions $f$ & $1\cdot 10^{-4}$ & \multirow{3}{*}{\shortstack{by half every \\ $5$ epochs}} & \multirow{3}{*}{$10^{-4}$} & \multirow{3}{*}{\shortstack{$21, 15, 10$ epochs \\ early stopping}} \\
         & add-on $1\!\!\times\!\!1$ convolution & $1\cdot 10^{-3}$ & & & \\
         & prototypical layer & $1\cdot 10^{-3}$ & &  & \\
    \hline
    \end{tabular}
    \caption{Learning schema for the ICICLE method.}
    \label{tab:learning_scheme}
\end{table*}

\section*{Results on Stanford Cars}

Table~\ref{table:cars_tasks} depicts how the ICICLE method works on the Stanford Cars dataset compared to other baseline methods. Results are consistent with the ones on CUB-200-2011 and show that ICICLE outperforms all standard CL learning methods adapted to ProtoPNet architecture. 

\begin{table*}[t]
\centering
\scriptsize 
\begin{tabular}{|c||c|c|c|c|c|c|}
\hline
& \multicolumn{3}{c|}{\textsc{Avg. inc. task-aware accuracy}} & \multicolumn{3}{c|}{\textsc{Avg. inc. task-task agnostic accuracy}} \\ [0.5ex]
\cline{2-7}
\textsc{Method} & \textsc{$4$ tasks} & \textsc{$7$ tasks} & \textsc{$14$ tasks} & \textsc{$4$ tasks} & \textsc{$7$ tasks} & \textsc{$14$ tasks}  \\ [0.5ex]
\hline
\hline
\textsc{Freezing} & $0.572 \pm 0.031$
& $0.518 \pm 0.041$ & $0.486 \pm 0.026$ & $0.309 \pm 0.012$ & $0.155 \pm 0.031$ & $0.092 \pm 0.014$ \\ [1ex]
\textsc{Finetuning} & $0.216 \pm 0.009$ & $0.167 \pm 0.011$ &
 $0.149 \pm 0.012$ & $0.182 \pm 0.006$ & $0.124 \pm 0.013$ & $0.057 \pm 0.001$ \\ [1ex]
\textsc{EWC} & $0.456 \pm 0.021$ & $0.315 \pm 0.037$ & $0.287 \pm 0.041$
& $0.258 \pm 0.019$ & $0.152 \pm 0.022$ & $0.011 \pm 0.009$ \\ [1ex]
\textsc{LWM} & $0.459 \pm 0.072$ & $0.416 \pm 0.048$ & $0.305 \pm 0.022$ 
& $0.233 \pm 0.026$ & $0.171 \pm 0.016$ & $0.080 \pm 0.008$ \\ [1ex]
\textsc{LWF} & $0.375 \pm 0.021$ & $0.356 \pm 0.024$  & $0.250 \pm 0.020$
&  $0.230 \pm 0.011$ & $0.171 \pm 0.005$ & $0.092 \pm 0.008$ \\ [1ex]
\hline
\textsc{ICICLE} & $\textbf{0.654} \pm \textbf{0.014}$ & $\textbf{0.645} \pm \textbf{0.003}$ & $\textbf{0.583} \pm \textbf{0.048}$
& $\textbf{0.335} \pm \textbf{0.005}$ & $\textbf{0.203} \pm \textbf{0.010}$ & $\textbf{0.116} \pm \textbf{0.018}$ \\[1ex]
\hline
\end{tabular}
\caption{Average incremental accuracy comparison for different numbers of tasks on Stanford Cars, demonstrating the negative impact of the high number of tasks to be learned on models' performance. Despite this trend, ICICLE outperforms the baseline methods across all task numbers.}
\label{table:cars_tasks}
\end{table*}

\section*{Comparison to GDumb}

Additionally, we compared our approach with GDumb~\cite{prabhu2020gdumb}, a baseline method in CL learning, in scenarios involving 3, 5, and 10 images per class with 4 tasks learning achieving 20.3, 34.2, 57.6, and 13.0, 26.7, 48.8 for task-aware and task-agnostic respectively. ICICLE outperformed GDumb with a small number of examples, and task-aware for GDumb-10 was the only exception where GDumb achieved a higher accuracy score.

\section*{Distant initialization}
In Table 5, we showed that proximity-based initialization is most beneficial. However, here in Figure~\ref{fig:distant}, we show how initialization of prototypical parts at a distance from already existing once generates concepts that are too general or carry information about background. 

\begin{figure}
    \centering
    \includegraphics[width=\linewidth]{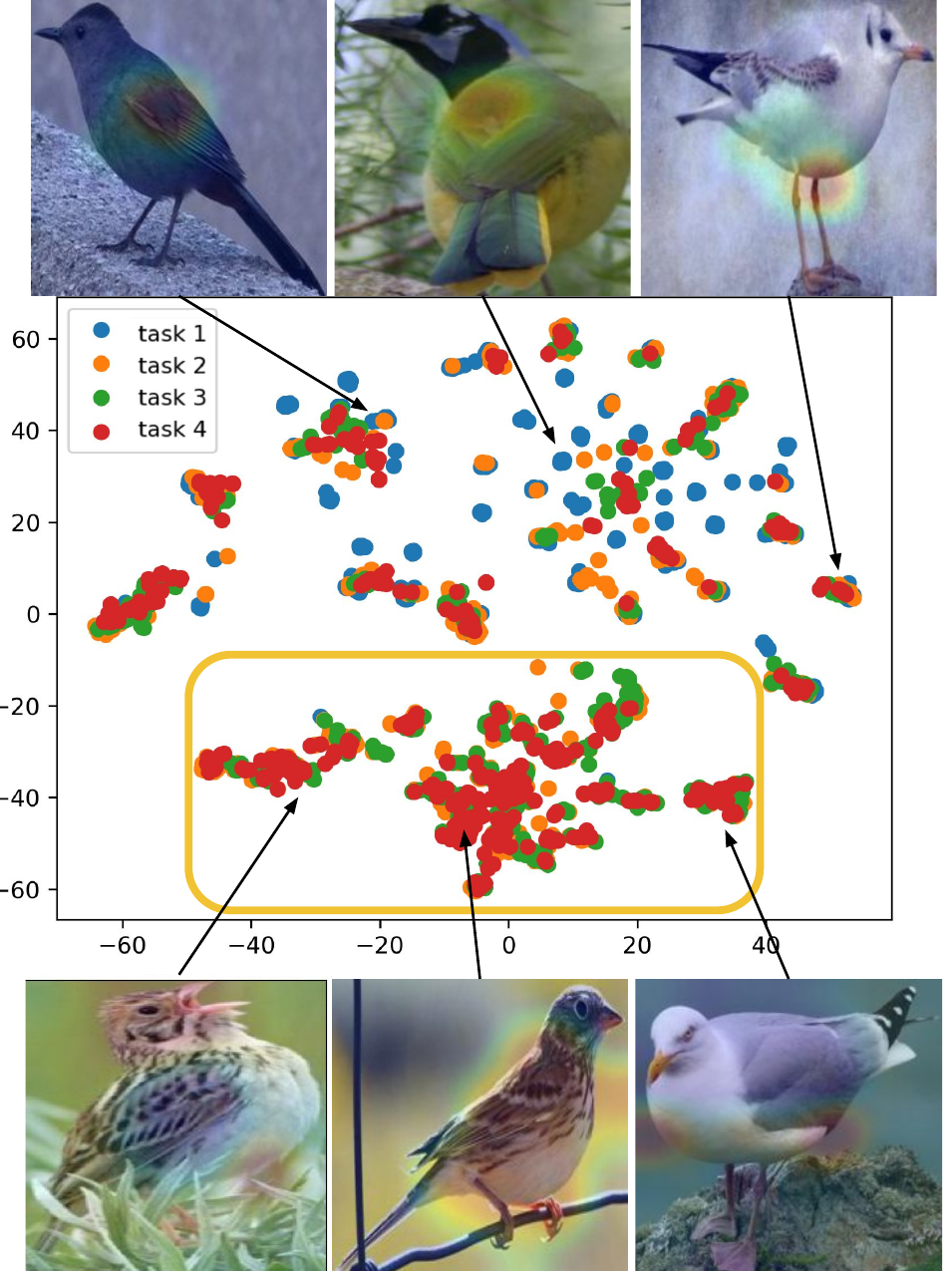}
    \caption{Image depicts the 2D TSNE projection of prototypes. One can observe that there is a cluster of prototypes from all tasks except the first one (yellow box). This is observable with distant initialization and prototype visualizations show that those prototypical parts are representing the background or vague concepts.}
    \label{fig:distant}
\end{figure}

\section*{Task-recency bias compensation}
Here, in Table~\ref{table:comp} we show what is the influence of task-recency bias compensation on each task accuracy for task agnostic scenario. After compensation, the accuracy on task 1 increased the most, but at the same time accuracy on all other tasks was sacrificed. However, the average accuracy after the compensation is increased. 

\begin{table}[t]
\centering
\scriptsize 
\begin{tabular}{|c||c|c|c|c|c|}
\hline
 & \textsc{task 1} & \textsc{task 2} & \textsc{task 3} & \textsc{task 4} & \textsc{Avg} \\ [0.5ex]
\hline
\hline
\textsc{Task aware} & $0.514 $
& $0.717 $ & $0.725 $ & $0.698 $ & $0.663 $ \\ [1ex]
\hline
\textsc{before} & $0.028$ & $0.301$ & $\textbf{0.434}$ & $\textbf{0.575}$ & $0.335$ \\ [0.25ex]  
\textsc{after} & $\textbf{0.233}$ & $\textbf{0.365}$ & $0.314$ & $0.486$ & $\textbf{0.350}$ \\ [0.25ex]
\hline
\end{tabular}
\caption{Task-recency bias compensation influence of a single run. Results show that our compensation method balance more the results per each task in task agnostic scenario.}
\label{table:comp}
\end{table}

\section*{Detailed results after learning each task}
In Table~\ref{table:4tasks}, and Table~\ref{table:10tasks} we show detailed accuracies of the ICICLE after learning each task for CUB-200-2011 on a single run with the same seed in four and ten tasks learning scenarios. 

\begin{table*}[t]
\centering
\scriptsize 
\begin{tabular}{|c||c|c|c|c|c|c|c|c|c|c||}
\hline
& \multicolumn{5}{c|}{\textsc{Task-aware accuracy}} & \multicolumn{5}{c||}{\textsc{Task-task agnostic accuracy}} \\ [0.5ex]
\cline{2-11}
 After& \textsc{task 1} & \textsc{task 2} & \textsc{task 3} & \textsc{task 4} & \textsc{Avg} & \textsc{task 1} & \textsc{task 2} & \textsc{task 3} & \textsc{task 4} & \textsc{Avg} \\ [0.5ex]
\hline
\textsc{task 1} & $0.806 $
& NA & NA & NA & $0.806 $  & $0.806 $
& NA & NA & NA & $0.806 $ \\ [1ex]
\textsc{task 2} & $0.740 $
& $0.759 $ & NA & NA & $0.750 $ & $0.089 $
& $0.747 $ & NA & NA & $0.418 $ \\ [1ex]
\textsc{task 3} & $0.622 $
& $0.736 $ & $0.759 $ & NA & $0.706 $ & $0.033 $
& $0.633 $ & $0.549 $ & NA & $0.404 $ \\ [1ex]
\textsc{task 4} & $0.514 $
& $0.717 $ & $0.725 $ & $0.698 $ & $0.663 $ & $0.028 $ & $0.484 $ & $0.378 $ & $0.505$ & $0.349 $ \\ [1ex]
\hline
\end{tabular}
\caption{Results of the ICICLE method before task-recency compensation for four task learning scenario after each learning episode.}
\label{table:4tasks}
\end{table*}

\begin{table*}[t]
\centering
\scriptsize 
\begin{tabular}{|c||c|c|c|c|c|c|c|c|c|c|c||}
\hline
& \multicolumn{11}{c||}{\textsc{Task-aware accuracy}}  \\ [0.5ex]
\cline{2-12}
 After& \textsc{task 1} & \textsc{task 2} & \textsc{task 3} & \textsc{task 4} & \textsc{task 5} & \textsc{task 6} & \textsc{task 7} & \textsc{task 8} &  \textsc{task 9} & \textsc{task 10} & \textsc{Avg} \\ 
\hline
\textsc{task 1} & $0.920 $
& NA & NA & NA  & NA & NA & NA & NA & NA &  NA & $0.920 $ \\ [1ex]
\textsc{task 2} & $0.666 $
& $0.869 $ & NA & NA  & NA & NA & NA & NA & NA &  NA & $0.767 $ \\ [1ex]
\textsc{task 3} & $0.462 $
& $0.818 $ & $0.858 $ & NA & NA  & NA & NA & NA & NA & NA & $0.713 $ \\ [1ex]
\textsc{task 4} & $0.420 $
& $0.751 $ & $0.774 $ & $0.774 $ & NA  & NA & NA & NA & NA & NA & $0.680$ \\ [1ex]
\textsc{task 5} & $0.314 $
& $0.625 $ & $0.680 $ & $0.672 $ & $0.784$  & NA & NA & NA & NA & NA & $0.615$ \\ [1ex]
\textsc{task 6} & $0.268 $
& $0.538 $ & $0.627 $ & $0.617 $ & $0.760$  & $0.747$ & NA & NA & NA & NA & $0.593$ \\ [1ex]
\textsc{task 7} & $0.265 $
& $0.476 $ & $0.584 $ & $0.617 $ & $0.713$  & $0.706$ & $0.769$ & NA & NA & NA & $0.590$ \\ [1ex]
\textsc{task 8} & $0.258 $
& $0.413 $ & $0.551 $ & $0.555 $ & $0.667$  & $0.634$ & $0.741$ & $0.764$ & NA & NA & $0.573$ \\ [1ex]
\textsc{task 9} & $0.253 $
& $0.398 $ & $0.494 $ & $0.492 $ & $0.598$  & $0.587$ & $0.701$ & $0.745$ & $0.852$ & NA & $0.569$ \\ [1ex]
\textsc{task 10} & $0.244 $
& $0.371 $ & $0.462 $ & $0.441 $ & $0.573$  & $0.560$ & $0.667$ & $0.729$ & $0.816$ & $0.803$ & $0.567$ \\ [1ex]
\hline
\hline
& \multicolumn{11}{c||}{\textsc{Task-agnostic accuracy}}  \\ [0.5ex]
\cline{2-12}
 & \textsc{task 1} & \textsc{task 2} & \textsc{task 3} & \textsc{task 4} & \textsc{task 5} & \textsc{task 6} & \textsc{task 7} & \textsc{task 8} &  \textsc{task 9} & \textsc{task 10} & \textsc{Avg} \\ 
\hline
\textsc{task 1} & $0.920 $
& NA & NA & NA  & NA & NA & NA & NA & NA &  NA & $0.920 $ \\ [1ex]
\textsc{task 2} & $0.010 $
& $0.869 $ & NA & NA  & NA & NA & NA & NA & NA &  NA & $0.439 $ \\ [1ex]
\textsc{task 3} & $0.0 $
& $0.060 $ & $0.854 $ & NA & NA  & NA & NA & NA & NA & NA & $0.305 $ \\ [1ex]
\textsc{task 4} & $0.0 $
& $0.0 $ & $0.339 $ & $0.751 $ & NA  & NA & NA & NA & NA & NA & $0.273$ \\ [1ex]
\textsc{task 5} & $0.0 $
& $0.0 $ & $0.030 $ & $0.323 $ & $0.746$  & NA & NA & NA & NA & NA & $0.220$ \\ [1ex]
\textsc{task 6} & $0.0 $
& $0.0 $ & $0.004 $ & $0.090 $ & $0.451$  & $0.684$ & NA & NA & NA & NA & $0.205$ \\ [1ex]
\textsc{task 7} & $0.0 $
& $0.0 $ & $0.0 $ & $0.020 $ & $0.193$  & $0.432$ & $0.712$ & NA & NA & NA & $0.194$ \\ [1ex]
\textsc{task 8} & $0.0 $
& $0.0 $ & $0.0 $ & $0.020 $ & $0.073$  & $0.233$ & $0.497$ & $0.643$ & NA & NA & $0.181$ \\ [1ex]
\textsc{task 9} & $0.0 $
& $0.0 $ & $0.0 $ & $0.0 $ & $0.035$  & $0.138$ & $0.338$ & $0.435$ & $0.676$ & NA & $0.180$ \\ [1ex]
\textsc{task 10} & $0.0 $
& $0.0 $ & $0.0 $ & $0.0 $ & $0.016$  & $0.070$ & $0.214$ & $0.285$ & $0.484$ & $0.643$ & $0.171$ \\ [1ex]
\hline
\end{tabular}
\caption{Results of the ICICLE method before task-recency compensation for ten task learning scenario after each learning episode.}
\label{table:10tasks}
\end{table*}

\section*{Analysis of the hyperparameters for baselines}
In Table~\ref{table:ewc}, Table~\ref{table:lwm}, and Table~\ref{table:lwf} we show the influence of the hyperparameters for each of the baseline methods, EWC, LWF, and LWM, respectively. Based on that, the parameters of these methods were chosen for comparison with ICICLE.

\begin{table*}[t]
\centering
\scriptsize 
\begin{tabular}{|c||c|c|c|c|c|c|c|c|c|c||}
\hline
& \multicolumn{5}{c|}{\textsc{Task-aware accuracy}} & \multicolumn{5}{c||}{\textsc{Task-task agnostic accuracy}} \\ [0.5ex]
\cline{2-11}
 $\alpha$ & 0.01 & 0.1 & 1.0 & 5.0 & 10.0 & 0.01 & 0.1 & 1.0 & 5.0 & 10.0 \\ [0.5ex]
\hline
  & $0.185 $
& $0.329$ & $0.441$ & $0.197$ & $0.167 $  & $0.170 $
& $0.185$ & $0.213$ & $0.168$ & $0.144 $ \\ [1ex]
\hline
\end{tabular}
\caption{Influence of the $alpha$ parameter in EWC on the accuracy of ProtoPNet architecture in four task learning scenario.}
\label{table:ewc}
\end{table*}

\begin{table*}[t]
\centering
\scriptsize 
\begin{tabular}{|c||c|c|c|c|c|c|c|c|c|c||}
\hline
& \multicolumn{5}{c|}{\textsc{Task-aware accuracy}} & \multicolumn{5}{c||}{\textsc{Task-task agnostic accuracy}} \\ [0.5ex]
\cline{2-11}
 $\gamma$ & 0.001 & 0.01 & 0.1 & 1.0 & 10.0 & 0.001 & 0.01 & 0.1 & 1.0 & 10.0 \\ [0.5ex]
\hline
  & $0.240$ & $0.240$ & $0.431$ & $0.355$ & $0.231$  
  & $0.209 $ & $0.209$ & $0.212$ & $0.209$ & $0.209$ \\ [1ex]
\hline
\end{tabular}
\caption{Influence of the $\gamma$ parameter in LWM on the accuracy of ProtoPNet architecture in four task learning scenario.}
\label{table:lwm}
\end{table*}

\begin{table*}[t]
\centering
\scriptsize 
\begin{tabular}{|c||c|c|c|c|c|c|c|c|c|c||}
\hline
& \multicolumn{5}{c|}{\textsc{Task-aware accuracy}} & \multicolumn{5}{c||}{\textsc{Task-task agnostic accuracy}} \\ [0.5ex]
\cline{2-11}
 $\lambda$ & 0.001 & 0.01 & 0.1 & 1.0 & 10.0 & 0.001 & 0.01 & 0.1 & 1.0 & 10.0 \\ [0.5ex]
\hline
  & $0.232$ & $0.232$ & $0.238$ & $0.359$ & $0.249 $  & 
  $0.207 $ & $0.210$ & $0.210$ & $0.231$ & $0.221 $ \\ [1ex]
\hline
\end{tabular}
\caption{Influence of the $\lambda$ parameter in LWF on the accuracy of ProtoPNet architecture in four task learning scenario.}
\label{table:lwf}
\end{table*}

\end{document}